\documentclass{article}
\usepackage{spconf,amsmath,graphicx}
\usepackage{bm}
\usepackage{graphicx}
\usepackage{multicol}
\usepackage[utf8]{inputenc}
\usepackage{amsmath,amssymb,amsthm,mathrsfs,amsfonts,dsfont}
\usepackage{booktabs}

\title{REVE: REGULARIZING DEEP LEARNING WITH VARIATIONAL ENTROPY BOUND}
\name{Antoine Saporta, Yifu Chen, Michael Blot, Matthieu Cord\sthanks{Thanks to ANR-15-CE23-0029-02 for funding.}}
\address{Sorbonne Université, CNRS, LIP6, 4 place Jussieu, 75005 Paris}
\begin{document}
%
\setlength{\abovedisplayskip}{1pt}
\setlength{\belowdisplayskip}{1pt}

\maketitle
\begin{abstract}
Studies on generalization performance of machine learning algorithms under the scope of information theory suggest that compressed representations can guarantee good generalization, inspiring many compression-based regularization methods. In this paper, we introduce \textsc{Reve}, a new regularization scheme. Noting that compressing the representation can be sub-optimal, our first contribution is to identify a variable that is directly responsible for the final prediction. Our method aims at compressing the class conditioned entropy of this latter variable. Second, we introduce a variational upper bound on this conditional entropy term. Finally, we propose a scheme to instantiate a tractable loss that is integrated within the training procedure of the neural network and demonstrate its efficiency on different neural networks and datasets. 
\end{abstract}

\newcommand{\Ys}{Z}
\newcommand{\ys}{z}
\newcommand{\YYs}{\mathcal{Z}}

\section{Introduction}
    Deep neural networks (DNNs) have demonstrated remarkable performance on image recognition and more particularly on image classification~\cite{alexnet,resnet} which is the main focus of this paper. 
    Despite constant progress of DNNs performances since \cite{alexnet}, their generalization ability is still largely misunderstood and theoretical tools such as PAC-based analysis seem limited to explain it, as demonstrated by \cite{rethinking}. Although regularization methods such as weight decay~\cite{weightdecay}, dropout~\cite{dropout} or batch normalization~\cite{batchnorm} are common practices to mitigate the ratio between the numbers of training samples and model parameters, the issue of DNNs regularization remains an open question.

    Recently, the information theory has been providing interesting tools to study DNNs performance from a new perspective \cite{soatto,IBaccuracy,nipsInformationGeneralization,bayes}. The objective of featuring is to create simple representations of the input while extracting enough knowledge about the class to predict in order to lower the complexity of the representation and achieve better generalization. 
    However, complexity may be given several interpretations in machine learning. 
    \cite{IB} introduces the Information Bottleneck (IB) framework that deals with an information theory-based notion of complexity.
    They state that a representation variable $Y$ that shares few information with the input $X$ can guarantee good generalization. 
    About applications on DNNs, many works investigate strategies consisting in optimizing a classification loss while compressing the representations. For example, \cite{IBvariational} develops a variational approximation of IB's criterion $I(Y,X)$. \textsc{Shade}~\cite{shade} implements a tractable estimation of another criterion based on conditional entropy $H(Y\mid C)$. The benefits of this criterion compared to $I(Y,X)$ are discussed in \cite{shade}: this conditional entropy allows to quantify the intra-class invariance of $Y$. This choice of criterion is validated by \textsc{Shade}'s excellent performance. 
    
    In this paper, we investigate a strategy following the same paradigm to optimize the DNN by defining a regularization loss that is added to the classification loss. Our contributions are threefold: first, we identify a variable $\Ys$ better suited for compression strategies. Second, as in \textsc{Shade}, we investigate a regularization based on conditional entropy, $H(\Ys \mid C)$, yet we develop an upper bound of this criterion in a general setup based on variational bounds on entropy measures. Third, we propose an instantiation protocol of our objective function applicable for DNN architectures and present extensive experiments on \textsc{Reve} to demonstrate its benefit for generalization.

\section{\textsc{Reve} Regularization}
    \label{sec:revecriterion}
    
    
    \subsection{\textsc{Reve} Variable}
        
        Suppose $Y$ is any encoding of the input $X$, possibly stochastic. If the decoder that maps $Y$ into the class space is linear, as most DDNs last layer, with corresponding multiplicative matrix $\bm{W}_d$, then it is possible to uniquely decompose $Y$ in the following way: $Y = \Ys + Y^{\ker}$, where $Y^{\ker}$ is the projection of $Y$ on the kernel of $\bm{W}_d$ ($\bm{W}_dY^{\ker}=0$) and $\Ys$ is the projection on the orthogonal complement of the kernel in the representation space. Thus $\bm{W}_d Y = \bm{W}_d\Ys$ and the term $Y^{\ker}$ is not taken into account in the prediction. Consequently, there is no point to compress it and we prefer to focus on the other part $\Ys$. We stress that the class space is generally of much lower dimension (the number of classes) than the input space and the representation space. Thus, the kernel of $\bm{W}_d$ is not reduced to zero and the kernel variable $Y^{ker}$ is very likely to contain a lot of information about the input $X$. 
        
        We study here a compression scheme on the part $\Ys$ of $Y$, the intermediate representation of the DNN, just before the last layer that maps the representation into the class space. 
        
        \begin{figure}[t]
        \centering
            \includegraphics[height=1.6cm,width=0.47\textwidth]{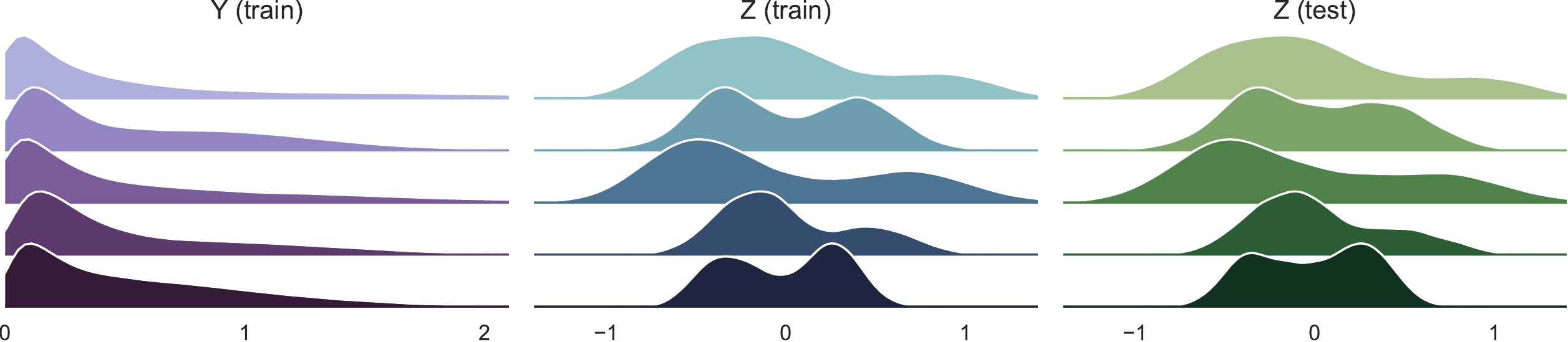}\\
            {\footnotesize(a) ResNet}\\

            \includegraphics[height=1.6cm,width=0.47\textwidth]{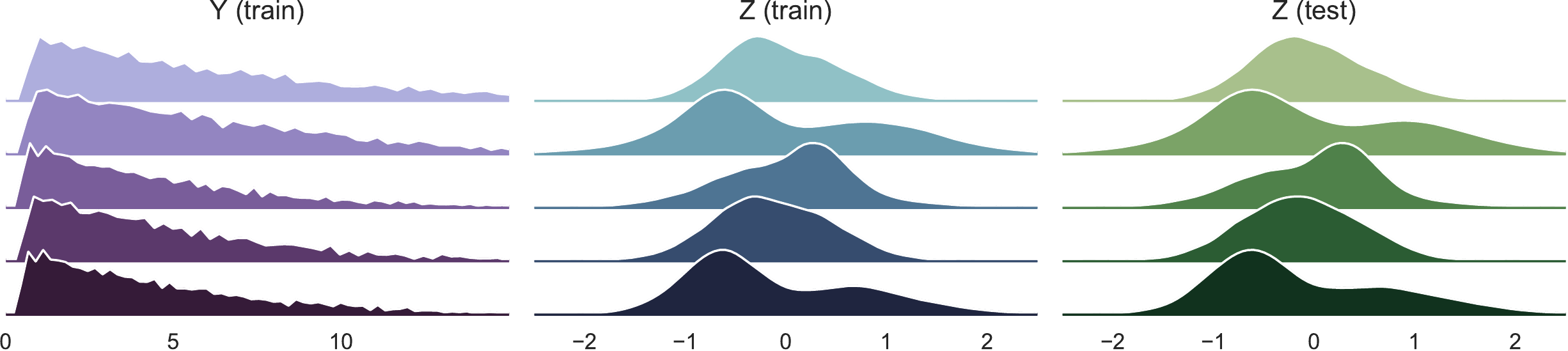}\\
            {\footnotesize(b) Alexnet}\\

        \caption{Kernel density estimates of five coordinates of $Y$ and $\Ys$ of baselines ResNet and Alexnet trained on CIFAR-10.}
        \label{fig:Zactivation}
        \end{figure} 
        
        As the informative part of the neuron activation, we expect $\Ys$ to exhibit interesting properties. As suggested in \cite{olah2017feature}, neurons may be interpreted as detectors of physical elements (object, shape, texture, \emph{etc.}) that are discriminative for the task. Figure~\ref{fig:Zactivation} shows the plotted estimations of the density for some coordinates of $\Ys$. 
        The bimodality that emerges for each neuron may corroborate this interpretation, each mode corresponding to the presence or absence of the pattern.
        
        

    \subsection{\textsc{Reve} Regularization Function}
        We adopt the conditional entropy $H(\Ys \mid C)$ proposed in \textsc{Shade}~\cite{shade} as regularization criterion, for which we develop a general variational upper bound. 
        First, notice that $H(\Ys\mid C)$ can be decomposed into simpler terms: $H(\Ys\mid C) = H(\Ys) - I(\Ys;C) = H(\Ys) - H(C) + H(C \mid \Ys)$. The entropy of the class $H(C)$ is entirely determined by the problem and is independent of any optimization. Therefore, we ignore it and define the following regularization criterion:
        \begin{equation}
            \label{eq:simplifiedEntropy}
            \mathcal{L}_\textsc{Reve}(\Ys,C) = H(\Ys) + H(C \mid \Ys).
        \end{equation}
        for $C \in \mathcal{C}$ and with any variable $\Ys\in \mathcal{\Ys}$ such that the quantities defined below exist. 
        The definitions of the two terms in $\mathcal{L}_\textsc{Reve}(\Ys,C)$ are as follow:
        \begin{align}
            H(\Ys) &= -\int\limits_{\mathcal{\Ys}} p(\bm{\ys})\log p(\bm{\ys}) \mathrm{d}\bm{\ys}\qquad\\
            H(C \mid \Ys) &= -\iint\limits_{\mathcal{\Ys}\:\mathcal{C}} p(\bm{\ys},\bm{c}) \log p(\bm{c}\mid \bm{\ys}) \mathrm{d}\bm{c} \mathrm{d}\bm{\ys} .
        \end{align}
            
        \emph{Variational Approximation:} For any space $\mathcal{Z}$, the Kullback-Leibler divergence between any two probability distributions $p(\bm{\ys})$ and $q(\bm{\ys})$ both defined on $\mathcal{\Ys}$ is always non-negative:
            
            $D_{KL}[p||q] = \int_{\mathcal{\Ys}} p(\bm{\ys})\log \frac{p(\bm{\ys})}{q(\bm{\ys})}\mathrm{d}\bm{\ys}  \ge 0$. 
            Thus if $p$ is the true distribution of a variable $\Ys$, any $q$ can be used to upper bound the entropy of $\Ys$:
            $ H(\Ys) \le - \int_{\mathcal{\Ys}} p(\bm{\ys})\log q(\bm{\ys})\mathrm{d}\bm{\ys}.$
            Moreover, there is equality if, and only if,  $p = q$ almost everywhere. This property ensures that the closer $p$ and $q$ are in the sense of the KL-divergence, the tighter our bound gets. Such a $q$ is said to be a \emph{variational approximation} of $p$.
        
            It is possible to upper bound the \textsc{Reve} criterion by defining variational approximations $q(\Ys)$ and $r(C \mid \Ys)$ of the true distributions $p(\Ys)$ and $p(C\mid \Ys)$, respectively:
            \begin{multline}
                \label{eq:globalBound}
                \mathcal{L}_\textsc{Reve}(\Ys,C) \le - \int\limits_\YYs p(\bm{\ys})\log q(\bm{\ys}) \mathrm{d}\bm{\ys}\\
                -\iint\limits_{\YYs\:\mathcal{C}}p(\bm{\ys},\bm{c})\log r(\bm{c}\mid\bm{\ys}) \mathrm{d}\bm{\ys}\mathrm{d}\bm{c}.
            \end{multline}
    
\section{Instantiating \textsc{Reve}}
    \label{sec:instantiate} 
    \begin{figure}[h]
        \centering
        \includegraphics[width=.48\textwidth]{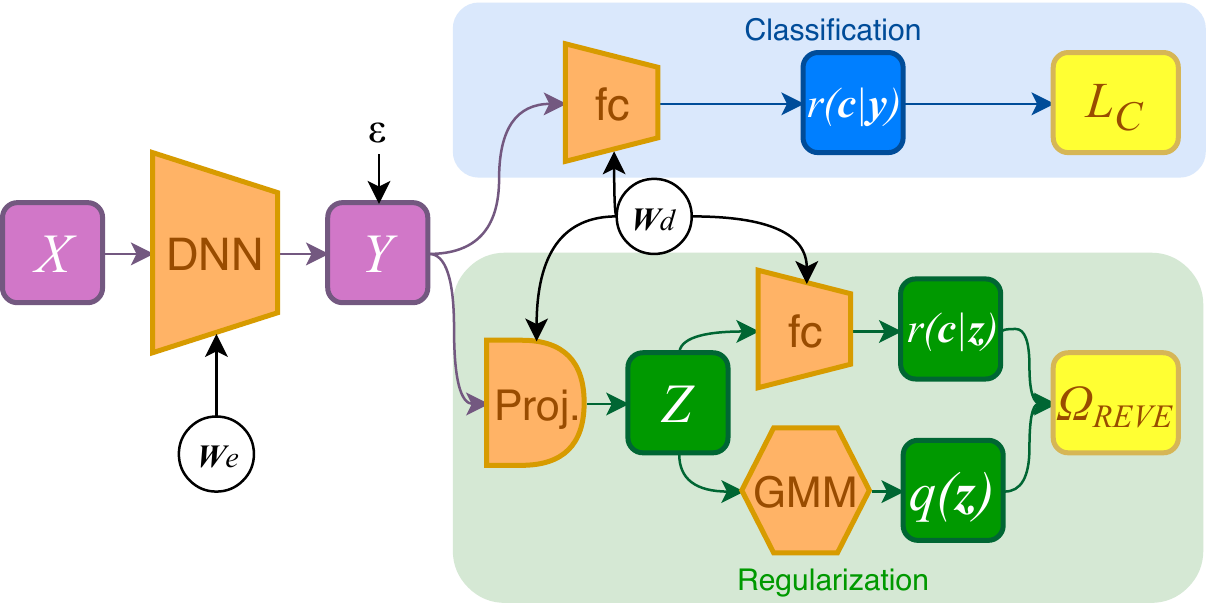}
        \caption{Overview of the \textsc{Reve} learning framework.
        }
        \label{fig:architecture}
    \end{figure}
    
    Similarly to \cite{IBvariational}, implementing \textsc{Reve} method necessitates to examine $Y\in \mathcal{Y}$ as a stochastic encoding of the input $X$. Its distribution follows the density $p(\bm{y}\mid \bm{x}) = p(\bm{y}\mid h(\bm{W}_e,\bm{x}))$ where $h(\bm{W}_e,\bm{x})$ is the output for input $\bm{x}$, of a neural network with parameters $\bm{W}_e$.
    We also introduce the decoder that maps $\mathcal{Y}$ onto the class probability space $\mathcal{P}(\mathcal{C})$: $\hat{C} = \text{softmax}(\bm{W}_dY + \bm{b})$ with $\bm{W}_d \in \mathds{R}^{|\mathcal{C}|\times \dim(\mathcal{Y}) }$ and $\bm{b} \in \mathds{R}^{|\mathcal{C}|}$. 
    
    When experimenting with deterministic baselines, we use $p(\bm{y}\mid \bm{x}) \sim \mathcal{N}(h(\bm{W}_e,\bm{x}), \sigma^2)$ as stochastic encoding, that is to say, $Y = h(\bm{W}_e, X) + \bm{\varepsilon}$ with $\bm{\varepsilon} \sim \mathcal{N}(0, \bm{\sigma}^2)$. Nevertheless, its expectancy is used to compute the prediction. Contrarily to \cite{IBvariational}, the variance $\bm{\sigma}^2$ is not learned but is a parameter of the method, effectively cutting the dimension of the output by a factor of two. Since $\bm{\varepsilon}$ is averaged during testing for computing the predicted class, the choice of its law is somewhat arbitrary. Thus, different values of $\bm{\sigma}^2$ 
    lead to different evaluations of entropy and only affect the regularization efficiency.

    \emph{Obtaining $\Ys$ from $Y$: }    
        The \emph{Singular-Value Decomposition} (SVD) is the natural factorization of a matrix for extracting both the kernel and its orthogonal complement. The compact SVD factorizes $\bm{W}_d$ as $\bm{U}\bm{\Sigma} \bm{V}^\top$, where $\bm{\Sigma}$ is a diagonal matrix of same rank $r$ as $\bm{W}_d$, containing all its non-zero singular values, and the $r$ columns of $\bm{U}$ (resp. $\bm{V}$) form an orthogonal set. Thus, $\bm{V}\bm{V}^\top$ is the orthogonal projection onto the orthogonal complement of the kernel and the part $\Ys$ is simply computed from $Y$ by: $\Ys = \bm{V}\bm{V}^\top Y$.

    \subsection{\textsc{Reve} Objective Function}
        We develop here the previous bound~\eqref{eq:globalBound}  by using a Monte Carlo method. 
        By applying the Bayes rule we get\footnote{We consider without further justifications that the conditions of the Fubini-Tonelli theorem allowing to invert the integrals are respected.}:
        \begin{multline}
            H(\Ys) \leq 
            -\iint\limits_{\mathcal{X}\:\YYs} p(\bm{x}) p(\bm{\ys}\mid \bm{x}) \log q(\bm{\ys}) \mathrm{d}\bm{x} \mathrm{d}\bm{\ys}.
            \label{finalentropyZbound}
        \end{multline}
        \begin{multline}
            H(C\mid \Ys) \leq\\ 
            -\iiint\limits_{\mathcal{X}\:\YYs\:\mathcal{C}} p(\bm{x}) p(\bm{c}\mid \bm{x}) p(\bm{\ys}\mid \bm{x}) \log r(\bm{c}\mid \bm{\ys}) \mathrm{d}\bm{x} \mathrm{d}\bm{\ys} \mathrm{d}\bm{c}.
        \label{finalmiZCbound}
        \end{multline}
        To obtain the result in line~\eqref{finalmiZCbound}, we exploit the Markov chain hypothesis $C \leftrightarrow X \leftrightarrow \Ys$ of the model: $p(\bm{c}\mid\bm{x}, \bm{\ys}) = p(\bm{c}\mid\bm{x})$ and thus $p(\bm{\ys},\bm{c}) = \int_\mathcal{X} p(\bm{x}) p(\bm{c}\mid \bm{x}) p(\bm{\ys}\mid \bm{x}) \mathrm{d}\bm{x}$.
        
        Using these results in \eqref{eq:simplifiedEntropy}, we obtain:
        \begin{multline}
           \mathcal{L}_\textsc{Reve}(\Ys, C) \leq - \iiint\limits_{\mathcal{X}\:\YYs\:\mathcal{C}} p(\bm{x}) p(\bm{c}\mid \bm{x}) p(\bm{\ys}\mid \bm{x}) \\\bigg(\log q(\bm{\ys}) + \log r(\bm{c}\mid \bm{\ys}) \bigg) \mathrm{d}\bm{x} \mathrm{d}\bm{\ys} \mathrm{d}\bm{c}.
        \label{finalbound}
        \end{multline}
        
        Then, we can use the empirical distribution of $(X,C)$ given by the mini-batch sample $(\bm{x}_n,\bm{c}_n)_{n=1..N}$ to use the Monte Carlo method. 
        The objective function can be written:
        \begin{multline}\label{exactloss}
            \Omega_{\textsc{Reve}}\big(X; C; \bm{W}_e; \bm{W}_d\big) = \\-\frac{1}{N}\sum\limits_{n=1}^N \left( \int\limits_\YYs p(\bm{\ys}\mid \bm{x}_n)\big(\log q(\bm{\ys}) + \log r(\bm{c}_n\mid \bm{\ys}) \big)\mathrm{d}\bm{\ys} \right)
        \end{multline}
        In order to estimate $\int\limits_\mathcal{\Ys}p(\bm{\ys}\mid \bm{x})\: \dots\: \mathrm{d}\bm{\ys}$, we again use the Monte Carlo method by sampling $\Ys$ from chosen probability $p(\bm{\ys} |\bm{x_n}) = p(\bm{\ys}\mid h(\bm{W}_e,\bm{x_n}))$. 
        $S$ samples $\bm{\varepsilon}_s$ of $\bm{\varepsilon}$ are drawn leading to the representation samples $\bm{y_{n,s}}= h(\bm{W}_e,\bm{x_n}) + \bm{\varepsilon}_s$ from which we deduce the samples $\bm{\ys_{n,s}}$ of $\Ys$. This approximation allows us to define our \textsc{Reve} loss function $\Omega_{\textsc{Reve}}$:
        
        \begin{multline}
            \Omega_{\textsc{Reve}}\big((\bm{x}_n,\bm{c}_n)_{n=1..N}; \bm{W}_e; \bm{W}_d\big) =\\ -\frac{1}{NS}\sum\limits_{n=1}^N\sum\limits_{s=1}^S \left( \log q(\bm{\ys}_{n,s}) 
            + \log r(\bm{c}_n\mid \bm{\ys}_{n,s}) \right).
        \label{approxloss}
        \end{multline}
        
        As suggested by Alemi et al.~\cite{IBvariational}, we choose to use $S=12$ which seemed to capture well enough the noisy distribution of $Y$ in our experiments.
        
        In everything that follows, we consider the approximation~\eqref{approxloss} as the objective function we want to minimize for regularization.
        
    \subsection{Model for $q$}
         
        Since the dimension of $\Ys$ can be arbitrarily large depending on the considered architecture, we start by adopting the \emph{mean field approximation}, considering the coordinates of $\Ys$ independent, in order to limit the complexity of our model:
        \begin{equation}
            q(\bm{\ys}) = \prod_{i=1}^{\dim(\mathcal{\Ys})} q_i(z_i)
        \label{meanfield}
        \end{equation}

       We have already mentioned in Section~\ref{sec:revecriterion} the intuition of a bimodal variable related to two modes, a neuron acting as a detector of a discriminative attribute. Thus, we choose to model $q_i(z_i)$ by a Gaussian Mixture Model (GMM) with two modes, $M_i=1$ and $M_i=0$:
        \begin{equation}
            q_i(z_i) = \alpha_i\mathcal{N}(z_i\mid \mu_{1,i},\sigma_{1,i}^2) + (1-\alpha_i)\mathcal{N}(z_i\mid \mu_{0,i}, \sigma_{0,i}^2)
        \label{bimodal}
        \end{equation}
        where the parameters $\bm{\alpha},\bm{\mu}_1, \bm{\sigma}_1^2,\bm{\mu}_0,\bm{\sigma}_0^2$ must be estimated.
        
        While \emph{Expectation-Maximization} is the standard algorithm for computing the parameters of a GMM, it is practically not applicable because the size of the mini-batches is in general too small. To circumvent this problem, we suggest to skip the E-step by directly approximating $\pi_i(z_i) =p(M_i=1\mid z_i)$ by the sigmoid function: $\pi_i(z_i) = \frac{1}{1+\exp(-z_i)}.$ Indeed, we observe empirically that one mode tends to be positive while the other one tends to be negative. Then, we compute the parameters of the two Gaussians on the mini-batch through the M-step. 
 
    \subsection{Model for $r$}
        Similarly to \cite{IBvariational}, the approximation $r(\bm{c}\mid \bm{\ys})$ can be defined by the multinomial logistic regression model at the end of the neural network using the Softmax $\mathcal{S}$:
        \begin{equation}
            r(\bm{c}\mid \bm{\ys}) =  \mathcal{S}(\bm{W}_d\bm{\ys}+\bm{b})_{\bm{c}} 
        \end{equation}
        where, for any $\bm{u}$, $u_{\bm{c}}$ is the coordinate of $\bm{u}$ related to the class $\bm{c}$.
        Unlike $q(\bm{\ys})$, the parameters of this variational approximation are defined by and optimized with the neural network.
        
    \subsection{\textsc{Reve} Regularization Loss}
        Combining previous results, the \textsc{Reve} loss is computed as:
        \begin{multline}
            \Omega_{\textsc{Reve}}\big((\bm{x}_n,\bm{c}_n)_{n=1..N}; \bm{W}_e; \bm{W}_d\big) = -\frac{1}{NS}\sum\limits_{n=1}^N\sum\limits_{s=1}^S \\\Bigg[
            \log \mathcal{S}(\bm{W}_d\bm{\ys}_{n,s}+\bm{b})_{\bm{c}_n}
            + \sum\limits_{i=1}^{\dim(\mathcal{\Ys})}\log \big( q_i(z_{n,s,i}) \big) \Bigg].
        \label{reveloss}
        \end{multline}
        where $q$ is defined in~\eqref{bimodal}. Figure~\ref{fig:architecture} summarizes the architecture of \textsc{Reve} and how it is embedded to a DNN.

\begin{table*}[tb]
        \centering
        \caption{Information-based regularization methods classification error (\%) results on CIFAR-10 and CIFAR-100 test sets. 
        }
        \begin{tabular}{l  c c c c@{\hskip 0.2in} c c c}
            \toprule
            && CIFAR-10 &&&& CIFAR-100 &\\
            \cmidrule{2-4} \cmidrule{6-8}
             & AlexNet & Inception Net & ResNet && AlexNet & Inception Net & ResNet\\
            \midrule
            Baseline & 15.62 & 6.10 & 4.08 && 48.29 & 27.36 & 20.70\\
            Dropout~\cite{dropout} & 12.63 & 6.04 & 3.93 && 41.32 & 27.26 & 20.16\\
            Information DO~\cite{infodropout} & 14.97 & 6.04 & NC && 47.97& 27.34& NC\\
            \textsc{Shade}~\cite{shade} + DO & 13.93 & 5.90 & 4.30&& 41.25 & 26.99 & 20.37\\
            \textsc{Reve} + DO & \textbf{12.54} & \textbf{5.78} &  \textbf{3.88} && \textbf{41.13} & \textbf{26.02} & \textbf{20.05}\\
            \bottomrule
        \end{tabular}
        \label{tab:cifar}
\end{table*}

\begin{table}[tb]
        \centering
        \caption{\textsc{Reve} Performance Analysis. Classification error (\%) results on CIFAR-10 and CIFAR-100 test sets. 
        }
        \begin{tabular}{@{\hskip 0.05in}l@{\hskip 0.1in}  c c@{\hskip 0in} c@{\hskip 0.1in} c c@{\hskip 0.05in}}
            \toprule
            & \multicolumn{2}{c}{CIFAR-10} && \multicolumn{2}{c}{CIFAR-100}\\
            \cmidrule{2-3} \cmidrule{5-6}
             & AlexNet & Inception  && AlexNet & Inception\\
            \midrule
            Baseline & 15.62 & 6.10 &&  48.29 & 27.36\\
            SGM \textsc{Reve}  & 14.24 & 6.17  && -  &  -\\
            KDE \textsc{Reve} & 13.86 & 6.04  && - & -\\
            \textsc{Reve} & 13.92 & 5.92 && 48.07 & 26.94\\
            \textsc{Reve} + DO & \textbf{12.54} & \textbf{5.78} && \textbf{41.13} & \textbf{26.02}\\
            \bottomrule
        \end{tabular}
        \label{tab:characteristics}
        \end{table}

\section{Experiments}

    \label{sec:experiments}
    We present experiments performed on several image databases: the CIFAR dataset~\cite{cifar} of animal and vehicle photos which includes color images of size $32 \times 32$ pixels classified in balanced and mutually exclusives classes, with $10$ and $100$ classes for CIFAR-10 and CIFAR-100, respectively; and the SVHN dataset~\cite{svhn} which includes color images of cropped digits of size $32\times 32$ pixels extracted from house number photos. 
    We experiment on three architectures, some near state-of-the-art on the CIFAR-10~\cite{cifar} dataset:\\
 \textbf{AlexNet}: an AlexNet-like model~\cite{alexnet} with three $5\times 5$ convolution + $2 \times 2$ max-pooling layers, then two fully connected layers with $1000$ neurons as intermediate representation;\\
 \textbf{Inception Net}: an Inception~\cite{inception} similar to the one in \cite{rethinking};\\
 \textbf{ResNet}: a ResNet architecture~\cite{resnet} similar to the one used in \cite{wideresnet} (depth $=28$ and $k=10$) and we mimic their training procedure, weight decay and learning rate schedule.
    
    Please note that we use our own implementation of these networks on TensorFlow~\cite{tensorflow}.
    
    When not stated otherwise: the networks are trained using SGD with $0.9$ momentum and an exponentially decaying learning rate; the batch size used is $128$; the cross entropy loss is used as classification loss and a weight decay of $10^{-3}$ is added to stabilize the training.
    Classic data augmentation techniques are used: the $32\times 32$ training images are padded to $40\times 40$ and randomly cropped to $32\times 32$; a random horizontal flip is applied. For preprocessing, we normalize the data using the channel means and standard deviations of the training set. The networks trained through this procedure without any additional regularization  constitute our baselines.
    
    The two hyperparameters $\bm{\sigma}^2$ and the Lagrange multiplier of the regularization loss $\beta_\textsc{Reve}$ are cross validated. Still, the regularization seems fairly stable to slight changes of these parameters ($\bm{\sigma}^2\sim 10^{-2}$ and $\beta_\textsc{Reve}\sim 10^{-4}$).

    \textbf{Ablation Study\quad}The results of some experiments around the characteristics of \textsc{Reve} can be found on Table~\ref{tab:characteristics}. 
        To better motivate our bimodal instantiation for $q(z)$, we experimented with a Single Gaussian Model (SGM \textsc{Reve}) and with a Kernel Density Estimator (KDE \textsc{Reve}). The SGM gives poor results in comparison to our bimodal model while the KDE requires more computation time for similar results. The proposed bound remains valid for any density model. Thus, the model can indeed be discussed and improved in order to get tighter bounds.  

       \noindent The addition of dropout improves the performance of \textsc{Reve}. 

    \textbf{Performance Results\quad}    Table~\ref{tab:cifar} compares the classification errors obtained on CIFAR with \textsc{Reve} to other regularization methods. For fair comparison, all experiments are done on the networks we implemented. \textsc{Reve} systematically gives the best performance of all these regularization methods.
    
    \noindent Table~\ref{tab:svhn} shows the results of similar experiments on the SVHN dataset which is significantly different from CIFAR. Yet, \textsc{Reve} still consistently improves the generalization performance of the networks. 
\begin{table}[tb]
            \centering
            \caption{Classification error (\%) results on SVHN test set. 
            }
            \begin{tabular}{l c c c}
                \toprule
                && SVHN &\\
                \cmidrule{2-4}
                 & AlexNet & Inception Net & ResNet\\
                \midrule
                Baseline & 7.68 & 3.78 & 3.40\\
                \textsc{Reve} & \textbf{6.55} & \textbf{3.29} & \textbf{3.11}\\
                \bottomrule
            \end{tabular}
        \label{tab:svhn}
        \end{table}

    \section{Conclusion}
    \label{sec:conclusion}    
        We have presented in this article a new regularization scheme, \textsc{Reve} aiming at minimizing an information-based criterion for compression: $H(\Ys\mid C)$. $\Ys$ is a variable extracted from the representation that only contains the information relevant for prediction. We develop a variational bound on this entropy and propose approximation models to instantiate it and regularize DNNs during SGD. Many experiments on three different datasets with various architectures demonstrate the benefits of \textsc{Reve} for generalization. Still, both theoretical and empirical analysis of our regularization leave the methods open to interpretation and enhancement. 

\bibliographystyle{IEEEbib}
\bibliography{main}

\begin{thebibliography}{10}

\bibitem{alexnet}
Alex Krizhevsky, Ilya Sutskever, and Geoffrey~E. Hinton,
\newblock ``Imagenet classification with deep convolutional neural networks,''
\newblock in {\em NIPS}, F.~Pereira, C.~J.~C. Burges, L.~Bottou, and K.~Q.
  Weinberger, Eds., 2012.

\bibitem{resnet}
Kaiming He, Xiangyu Zhang, Shaoqing Ren, and Jian Sun,
\newblock ``Deep residual learning for image recognition,''
\newblock in {\em CVPR}, 2016.

\bibitem{rethinking}
Chiyuan Zhang, Samy Bengio, Moritz Hardt, Benjamin Recht, and Oriol Vinyals,
\newblock ``Understanding deep learning requires rethinking generalization,''
\newblock {\em ICLR}, 2017.

\bibitem{weightdecay}
Anders Krogh and John~A. Hertz,
\newblock ``A simple weight decay can improve generalization,''
\newblock in {\em NIPS}, 1992.

\bibitem{dropout}
Nitish Srivastava, Geoffrey Hinton, Alex Krizhevsky, Ilya Sutskever, and Ruslan
  Salakhutdinov,
\newblock ``Dropout: A simple way to prevent neural networks from
  overfitting,''
\newblock {\em JMLR}, 2014.

\bibitem{batchnorm}
Sergey Ioffe and Christian Szegedy,
\newblock ``Batch normalization: Accelerating deep network training by reducing
  internal covariate shift,''
\newblock {\em JMLR}, 2016.

\bibitem{soatto}
Alessandro Achille and Stefano Soatto,
\newblock ``Emergence of invariance and disentanglement in deep
  representations,''
\newblock in {\em ArXiv}, 2017.

\bibitem{IBaccuracy}
Ran Gilad-Bachrach, Amir Navot, and Naftali Tishby,
\newblock ``An information theoretic tradeoff between complexity and
  accuracy,''
\newblock in {\em Learning Theory and Kernel Machines}. Springer, 2003.

\bibitem{nipsInformationGeneralization}
Xu~Aolin and Raginsky Maxim,
\newblock ``Information-theoretic analysis of generalization capability of
  learning algorithms,''
\newblock in {\em Advances in Neural Information Processing Systems 30},
  I.~Guyon, U.~V. Luxburg, S.~Bengio, H.~Wallach, R.~Fergus, S.~Vishwanathan,
  and R.~Garnett, Eds. 2017, pp. 2524--2533, Curran Associates, Inc.

\bibitem{bayes}
Yarin Gal and Zoubin Ghahramani,
\newblock ``Dropout as a bayesian approximation: Representing model uncertainty
  in deep learning,''
\newblock in {\em ICML}, 2017.

\bibitem{IB}
N.~Tishby, F.~C. Pereira, and W.~Bialek,
\newblock ``The information bottleneck method,''
\newblock {\em Annual Allerton Conference on Communication, Control and
  Computing}, 1999.

\bibitem{IBvariational}
A.~A. Alemi, I.~Fischer, J.~V Dillon, and K.~Murphy,
\newblock ``Deep variational information bottleneck,''
\newblock in {\em ICLR}, 2017.

\bibitem{shade}
Michael Blot, Thomas Robert, Nicolas Thome, and Matthieu Cord,
\newblock ``{SHADE}: {SHA}nnon {DE}cay information-based regularization for
  deep learning,''
\newblock in {\em IEEE International Conference on Image Processing (ICIP)},
  2018.

\bibitem{olah2017feature}
Chris Olah, Alexander Mordvintsev, and Ludwig Schubert,
\newblock ``Feature visualization,''
\newblock {\em Distill}, 2017,
\newblock https://distill.pub/2017/feature-visualization.

\bibitem{infodropout}
A.~Achille and S.~Soatto,
\newblock ``Information dropout: learning optimal representations through noisy
  computation,''
\newblock {\em ArXiv}, 2016.

\bibitem{cifar}
A.~Krizhevsky,
\newblock {\em Learning multiple layers of features from tiny images},
\newblock Ph.D. thesis, Computer Science Department University of Toronto,
  2009.

\bibitem{svhn}
Adam Coates Alessandro Bissacco Bo Wu Andrew Y.~Ng Yuval~Netzer, Tao~Wang,
\newblock ``Reading digits in natural images with unsupervised feature
  learning,''
\newblock in {\em NIPS Workshop on Deep Learning and Unsupervised Feature
  Learning}, 2011.

\bibitem{inception}
Christian Szegedy, Wei Liu, Yangqing Jia, Pierre Sermanet, Scott~E. Reed,
  Dragomir Anguelov, Dumitru Erhan, Vincent Vanhoucke, and Andrew Rabinovich,
\newblock ``Going deeper with convolutions,''
\newblock {\em CoRR}, vol. abs/1409.4842, 2014.

\bibitem{wideresnet}
Sergey Zagoruyko and Nikos Komodakis,
\newblock ``Wide residual networks,''
\newblock in {\em BMVC}, 2016.

\bibitem{tensorflow}
Mart\'{\i}n Abadi, Paul Barham, Jianmin Chen, Zhifeng Chen, Andy Davis, Jeffrey
  Dean, Matthieu Devin, Sanjay Ghemawat, Geoffrey Irving, Michael Isard,
  Manjunath Kudlur, Josh Levenberg, Rajat Monga, Sherry Moore, Derek~G. Murray,
  Benoit Steiner, Paul Tucker, Vijay Vasudevan, Pete Warden, Martin Wicke, Yuan
  Yu, and Xiaoqiang Zheng,
\newblock ``Tensorflow: A system for large-scale machine learning,''
\newblock in {\em Proceedings of the 12th USENIX Conference on Operating
  Systems Design and Implementation}, Berkeley, CA, USA, 2016, OSDI'16, pp.
  265--283, USENIX Association.

\end{thebibliography}

\end{document}